# A GUIDE TO BAYESIAN NETWORKS SOFTWARE PACKAGES FOR STRUCTURE AND PARAMETER LEARNING - 2025 EDITION

**Joverlyn Gaudillo[1], Nicole Astrologo[1], Fabio Stella[2], Enzo Acerbi[1] and Francesco Canonaco[1,2]**
[1]Minutia.AI Pte. Ltd., Singapore
[2] Department of Informatics, Systems and Communication,
University of Milano-Bicocca, Milano, Italy
[*]Corresponding author: francesco.canonaco@minutia.ai

## ABSTRACT

A representation of the cause-effect mechanism is needed to enable artificial intelligence to represent how the world works. Bayesian Networks (BNs) have proven to be an effective and versatile tool for this task. BNs require constructing a structure of dependencies among variables and learning the parameters that govern these relationships. These tasks, referred to as structural learning and parameter learning, are actively investigated by the research community, with several algorithms proposed and no single method having established itself as standard. A wide range of software, tools, and packages have been developed for BNs analysis and made available to academic researchers and industry practitioners. As a consequence of having no one-size-fits-all solution, moving the first practical steps and getting oriented into this field is proving to be challenging to outsiders and beginners. In this paper, we review the most relevant tools and software for BNs structural and parameter learning to date, providing our subjective recommendations directed to an audience of beginners. In addition, we provide an extensive easy-to-consult overview table summarizing all software packages and their main features. By improving the reader's understanding of which available software might best suit their needs, we improve accessibility to the field and make it easier for beginners to take their first step into it.

## 1 Introduction

Bayesian networks (BNs) have established themselves over the years as a powerful framework for modeling and analyzing complex systems under conditions of uncertainty. BNs represent probabilistic relationships among variables in a graphical way that allows efficient inference and intuitive causal reasoning when specific assumptions are met. A BN is a probabilistic graphical model consisting of a directed acyclic graph (DAG) where nodes represent variables and edges represent probabilistic dependencies. Each node in the graph is associated with a conditional probability distribution that quantifies the relationship between that node and its parent nodes (i.e., the nodes that have a direct effect on it) [1]. BNs enable the modeling of multifactorial systems by capturing both direct and indirect dependencies among variables, facilitating probabilistic reasoning and decision-making.

Learning the DAG of a BN from data is a foundational step of the model construction process. Identifying potentially causal relationships among variables can uncover insights into the underlying mechanisms governing a system. For this purpose, a multitude of algorithms have been developed over the years; these methods are typically categorized into four groups: constraint-based, score-based, functional, and gradient-based. An overview of structure learning approaches is beyond the scope of this document; a comprehensive assessment of state-of-the-art methodologies can be found in [2, 3, 4, 5]. Parameter learning is another critical task in BNs development. Given the DAG, the objective of parameter learning is to estimate the parameters of the conditional probability distributions associated with each node, which is essential for inference and prediction. For a comprehensive review of parameter learning strategies, challenges, and algorithms, refer to the works of [6, 7].

Approaching the study of BN framework requires a solid understanding of fundamental principles in disciplines such as probability and computer science. Assuming that the reader is already familiar with these foundations, some remarkable readings on causality and BNs science are offered by Probabilistic Graphical Models Principles and Techniques [1], Bayesian Artificial Intelligence [8], Probabilistic Reasoning in Intelligent Systems [9], Bayesian Networks with Examples in R [10], Bayesian Networks in R with Application in the field of System Biology[11], Bayesian Networks and Influence Diagrams [12]. This document assumes that the reader is equipped with the necessary foundational knowledge and is ready to engage in practical hands-on work.

Over the past five years, the field of causality and BNs development has seen an influx of numerous packages with no single solution being able to cater to all requirements and scenarios; this abundance of options is often challenging for individuals trying to gain hands-on experience with BNs. This document simplifies structure and parameter learning in BNs by providing a comprehensive overview of available software packages. In addition, we offer our subjective recommendations on selecting the best tools based on the reader's specific objectives. The paper is structured as follows: Section 1 provides a systematic review of both open-source and commercial software. Section 2 offers guidance on selecting tools suitable for beginners. Section 3 summarizes the key contributions of this work. Table 1 includes a concise summary table for quick reference to all reviewed tools.

## 2 Software Tools and Packages

### 2.1 gCastle

gCastle [13] is an end-to-end Python toolbox created by Huawei Noah's Ark Lab for causal structure learning. The package is equipped with functionalities such as data generation from simulated or real-world datasets, causal structure learning, and evaluation metrics.

### 2.2 bnlearn

bnlearn [14] is an R package developed by Marco Scutari and first released in 2007 with functionality to learn the structure of BNs, parameter estimation, and inference. After 10 years of continuous development, the package has grown to accommodate a multitude of algorithms from the literature. The package implements constraint-based algorithms, e.g., Peter-Clark (PC), Grow-Shrink (GS), Incremental Association Markov Blanket (IAMB), Inter-IAMB, Fast-IAMB, IAM-False Discovery Rate (FDR), Semi-Interleaved HITON-PC, and Max-Min Parents and Children (MMPC), pairwise-based algorithms, e.g., Algorithm for the Reconstruction of Accurate Cellular Networks (ARACNe) and Chow-Liu (ARACNE and Chow-Liu), score-based, e.g., Hill-Climbing (HC) and Tabu Search, hybrid algorithms, e.g., Hybrid Parents and Children (HPC), Max-Min HC (MMHC), Restricted Structural Maximum Algorithm 2 (RSMAX2), and Tree-augmented Naive Bayes (TAN), structure learning algorithms for discrete, Gaussian and conditional Gaussian networks, along with many score functions and conditional independence tests. Some utility functions (model comparison and manipulation, random data generation, arc orientation testing, simple and advanced plots) are included, as well as support for parameter estimation, e.g., maximum likelihood estimation (MLE) and Bayesian estimation, and inference, conditional probability queries, cross-validation, bootstrap, and model averaging.

### 2.3 pgmpy

Pgmpy [15] is a Python library developed in 2015 by Ankur Ankan to work with probabilistic graphical models. It allows users to create their graphical models and then perform inferences or map queries to them. The library implements several inference algorithms like variable elimination, belief propagation, etc. The library is designed with a modular structure, allowing users to access dedicated classes for commonly used graphical models like Naive Bayes (NB) and hidden Markov models, eliminating the need to build them from base models. Currently, it includes implementations of various algorithms for structure learning, parameter estimation, both approximate, i.e., sampling-based, and exact inference, as well as causal inference.

### 2.4 Tetrad

Tetrad [16] is a Java suite of software for the discovery, estimation, and simulation of causal models developed by the Carnegie Mellon University-Causal Learning and Reasoning (CMU-CLeaR) group. Some of its basic features for beginners include the ability to load existing datasets, load existing causal graphs, and create a new causal graph. For practitioners, the tool is equipped with advanced functionalities, such as specifying prior knowledge on constraint-based algorithms, manipulating data by imputing missing values, discretizing data, simulating data from statistical models, and computing the probability distribution of any variable, among others. It features a graphical user interface (GUI) and offers popular constraint-based algorithms for causal discovery such as PC, Fast Causal Inference (FCI), PC-Max, Conservative PC (CPC), and MLE for parameter learning.

### 2.5 Causal Command (CMD)

Causal-cmd is a Java application that offers a command-line interface tool for causal discovery algorithms developed by the Center for Causal Discovery [17]. Currently, the application includes more than 30 algorithms for causal discovery.

### 2.6 causal-learn

Causal-learn [18] is a Python translation and extension of the Tetrad Java code (refer to the Tetrad package) developed by CMU-CLeaR group [19]. It offers implementations of up-to-date causal discovery methods, as well as simple and intuitive Application Programming Interfaces (APIs). The project is in development; for more details, please refer to the repository [20].

### 2.7 pcalg

Pcalg [21] is an R package developed by Markus Kalisch *et al.* in 2006. It offers constraint-based algorithms such as PC, FCI, and Really FCI (RFCI) as well as hybrid and score-based algorithms for causal discovery. For a complete description of the methodologies offered by the package, refer to [22].

### 2.8 LiNGAM

Linear Non-Gaussian Acyclic Model (LiNGAM) [23] is a Python package for causal discovery developed by T. Ikeuchi *et al.* The package offers many causal discovery algorithms for linear non-Gaussian models such as Direct-LiNGAM, Linear Non-Gaussian Models for Latent Factors (LiNA), and Vector Autoregressive Models-LiNGAM (VAR-LiNGAM).

### 2.9 CDT

CDT [24] is a Python package for causal inference in graphical models and pairwise settings (compatible with Python $\geq$ 3.5). Developed by Diviyan Kalainathan and Olivier Goudet, CDT provides tools for structure learning and dependency analysis. It leverages on NumPy, scikit-learn, PyTorch, and R to implement various algorithms for causal discovery, including methods from bnlearn and pcalg. The package is particularly suited for analyzing observational data, offering both classical and deep learning-based approaches to causal structure recovery.

### 2.10 pyAgrum

pyAgrum [25] is a Python wrapper for the C++ aGrUM library [26]. It offers a high-level interface to aGrUM, enabling users to create, model, learn, apply, compute, and integrate BNs and other graphical models. Some specific (Python and C++) codes are added to simplify and extend the aGrUM API. The package contains causal discovery, parameter learning, and inference algorithms.

### 2.11 bnlearn (Python)

Bnlearn [27] is a Python package for causal discovery, parameter learning and inference developed by Erdogan Taskesen. It implements the most classical approaches for causal discovery such as HC, exhaustive search, Chow-Liu, TAN, PC, and MLE, as well as Bayesian estimation for parameter learning.

### 2.12 OpenMarkov

OpenMarkov [28] is a Java open-source software tool developed by the Research Centre for Intelligent Decision-Support Systems [29]. OpenMarkov comes with a user interface and can perform causal discovery employing the PC algorithm and HC search.

### 2.13 pomegranate

Pomegranate [30], a Python package developed by Jacob Schreiber, offers efficient and versatile probabilistic models, spanning from individual probability distributions to composite models including BNs and hidden Markov models. The package offers both constraint-based and score-based algorithms, as well as parameter learning procedures.

### 2.14 BayesFusion

BayesFusion [31] is a commercial software offering different solutions for causal discovery, parameter learning, and inference. Their flagship product is GeNIe, a tool for artificial intelligence and machine learning that has at its core the BN framework and other types of graphical probabilistic models. The SMILE engine allows the user to include custom applications that can be written in a variety of programming languages, e.g., C++, Python, Java, .NET, R, Matlab. Models created with GeNIe or SMILE can be shared or used on mobile devices via BayesMobile, or through a web browser with BayesBox.

### 2.15 BayesiaLab

BayesiaLab [32] is a commercial software developed by Dr. Lionel Jouffe and Dr. Paul Munteanu and their team. It offers plenty of algorithms for causal discovery, parameter learning, and inference. The software includes a graphical user interface and is well documented.

### 2.16 Bayes Server

Bayes Server [33] is a commercial software developed by Bayes Server Ltd. Besides the most well-known algorithms for causal discovery, parameter learning and inference, the software offers a wide range of tools for diagnostic, anomaly detection and decision-making under uncertainty which have at their core the BN framework. Bayes Server can be used in the cloud as well as on a local machine through a GUI. It offers an advanced user interface accessible programmatically via a number of APIs that can be used via Java, Matlab, Python, Spark and R.

## 3 My Causal Path: Picking the Right Tool as a Beginner

This section aims to assist beginners select the ideal package or software that best suits their needs. The first subsection focuses on causal discovery tools, while the second presents tools that support functionalities for both parameter learning and causal discovery are presented. Finally, the last subsection discusses commercial software that offers additional features such as optimized user-interfaces and professional customer support. Note that while the previous section provided a comprehensive overview of available solutions, this section shortlists and discusses only those we consider most suitable for beginners.

### 3.1 Tools for Causal Discovery Only

When the goal is limited to learning from data the structure representing the dependencies among variables, gCastle, CDT, and LiNGAM are three tools that represent viable solutions and provide easy access to those functionalities. In particular, gCastle by Huawei Noah's Ark Lab is in our opinion one of the most accessible and comprehensive causal discovery open-source Python libraries at the time of writing this document. It offers various cutting-edge approaches for recovering the structure of causal networks ranging from score-based to gradient-based and hybrid algorithms. For each algorithm, the documentation [34] offers a detailed practical example, making the tool very friendly to beginners. Various examples can also be found in Causal Inference and Discovery in Python (Part 3: Causal Discovery) [35], which offers the user the ability to dive deeper into any particular functionality offered by the tool. Moreover, gCastle can also be used via a GUI available at [36], which provides a friendlier version of the interface that does not involve coding. CDT is another great package that we feel confident in recommending. Its documentation contains several examples that will guide users step-by-step into their first structural learning attempts [37]. CDT has the largest collection of algorithms for causal discovery among all the other reviewed tools for beginners, some of which can be run using Pytorch as well [38].

For time series data, LiNGAM models provide a useful approximation of a stationary AR(p) process [39]. The LiNGAM package offers a comprehensive set of algorithms for learning linear non-Gaussian models, assuming non-Gaussian continuous error variables (with at most one exception). Its documentation [40] includes both theoretical overviews and practical code examples for each model.

When the goal is performing causal discovery on big data, Causal-Command represents a valid option. This Java library implements several algorithms for causal discovery and can be used via shell script or as part of a Java-based application. We perceive this library to be less user-friendly compared to the ones mentioned above, thus, we deem Causal-Command a good fit for more intermediate or advanced users. A short document is available [41] presenting the main functionalities and commands.

To conclude our assessment of tools specialized in structural learning, we consider CDT to be the best choice when having a large set of available methodologies is desirable. For example, CDT could be the most useful for training or educational purposes, where assessing and comparing the effectiveness of various methods is needed. CDT is also the best choice when an interface with Pytorch is required or preferred. On the other hand, gCastle is not as rich in causal discovery algorithms as CDT, but it couples a friendly code-free user interface with highly curated and rich documentation. Despite both CDT and gCastle being capable of handling linear non-Gaussian models, LiNGAM is a dedicated library for this specific model type, making LiNGAM the preferred tool in such scenarios.

### 3.2 Tools for Causal Discovery & Parameter Learning

In most cases, one may want to learn not only the structure of the causal network from the data but also the associated parameters. To this end, several tools extensively cover both functional areas while offering great simplicity of use. One of the most complete and well-maintained tools to date is bnlearn. Apart from the remarkable availability of built-in methods for parameter learning, structural learning, inference, missing data handling, and model validation strategies, what makes bnlearn stand out is its documentation and practical examples [42]. Remarkably, most methods and examples are thoroughly explained in the books Bayesian Networks in R and Bayesian Networks With Examples in R [10], of which the creator of bnlearn is co-author.

A valid alternative to bnlearn is represented by pgmpy. Unlike bnlearn, which provides methods for the static scenario only, pgmpy partially covers the dynamic case as well. This is an important feature given that a great portion of real-world problems and systems include time-dependent components. On the other hand, pgmpy's methodologies are comparatively less rich than those of bnlearn, particularly for the structural learning task. Nonetheless, pgmpy compensates for this limitation by offering more comprehensive documentation. Abundant examples are available in the practical notebooks [43] section, along with tutorial notebooks [44], both of which are beneficial for moving the first steps into this field.

Another alternative to bnlearn is pyAgrum. Just like pgmpy, pyAgrum provides methods for static and dynamic scenarios, making it a valid option for time-dependent real-world problems. pyAgrum offers comprehensive documentation including tutorials [45], examples [46] and applications with interactive widgets [47]. An important resource offered by pyAgrum is a list of implemented solutions [48] to the problems presented in the 'Book of Why' by Judea Pearl. PyAgrum not only provides rich and well-organized documentation, but also offers a wide array of causal discovery methodologies. For example, it implements greedy hill climbing (GHC), local search with tabu-list (LS-TL), Multivariate Information-based Inductive Causation (MIIC), Chow-Liu, NB, TAN, and K2 algorithms.

A less sophisticated yet relevant package is the Python version of the original bnlearn (which is an R package). Although it is not as rich in methodologies as Pgmpy and the original bnlearn (only a handful of causal discovery algorithms are available in it), the Python version of bnlearn offers an intuitive interface and its documentation is as rich and well-curated as the original R version [49]. The documentation not only presents many code snippets followed by the associated output but also provides a brief introduction to the theory behind it.

In conclusion, for those who are familiar with R, bnlearn represents the best choice, especially when coupled with the aforementioned books. For practitioners who prefer Python and/or need to model dynamic systems, pgmpy and pyAgrum are the best alternatives to bnlearn; the multitude of examples contained in pgmpy and pyAgrum documentation provides tremendous added value for beginners and/or practitioners moving their first steps in this field. The Python version of bnlearn offers a more straightforward interface than the other options; however, it does come with a limited number of structure learning algorithms, making it suitable for readers seeking to begin with simpler implementations.

### 3.3 Commercial Software

For a wider and more flexible application of BNs frameworks in industry settings where cloud computing might be involved, the resulting models often need to be shared and accessed from a variety of devices, including mobile devices, where no-code solutions may be preferable. In addition, in these kind of scenarios, professional support is usually needed, making the open-source packages described in the previous sections unsuitable. In this section, we illustrate some practical commercial solutions that might satisfy the needs of larger industry organizations.

For this purpose, Bayes Server would be our recommended choice; a demo can be found at the following link [50]. Bayes Server offers comprehensive documentation with several examples of how to interact with the GUI [51]. The code section of the documentation [52] acts as a central repository for practical examples of how to work with the Bayes Server API, while in the solution areas [53], many real use cases from different domains ranging from aerospace to healthcare are presented and described in detail. Bayes Server offers a commercial and academic license [54].

GeNIe by BayesFusion LLC is a valid alternative to Bayes Server. GeNIe makes use of the SMILE engine, a library of C++ classes that implement causal and parameter learning, as well as inference, which can be called via API. SMILE can be used via Java, Python, R, and .NET using the following wrappers: jSMILE (Java and environments that can instantiate and use the JVM), PySMILE (Python 2.7 and 3.x), rSMILE (R 3.x), SMILE.NET (.NET). Another component of GeNIe is BayesBox, an interactive repository where graphical models can be uploaded, shared, and consulted from a variety of devices, including mobiles.

A demo of BayesBox can be found here [55]. BayesFusion also offers detailed documentation [56], which contains information about GeNIe and its main features, as well as examples and introductory materials for SMILE. The forum [56] is also well-populated and can be a useful resource for support.

A viable alternative to BayesServer and GeNIe is BayesiaLab. BayesiaLab has a commercial license and offers an intuitive GUI, APIs [57], and many useful resources such as an ebook [58] that includes several tutorials. Webinars, tutorials, and use cases that will help users navigate the multitude of features offered by BayesiaLab are also available [59]. It is worthwhile to mention that the BayesiaLab API framework can be accessed using Java only.

In conclusion, both BayesServer and GeNIe can suit the aforementioned contexts. They are both equipped with a web platform that features a user-friendly interface and ready-to-use examples, and both software can be used on mobile devices. For BayesServer and GeNIe, pricing and licensing models can be the deciding factors in determining which tool best suits the reader's needs after having tried their trial and demo versions. This might not apply to BayesiaLab, as users cannot try the software on the website before purchasing it. Additionally, BayesiaLab can only be used with Java.

## 4 Conclusion

This paper presented an overview of the most recent tools and software for BNs causal discovery and parameter learning. The tools were reviewed from the perspective of a beginner seeking to gain hands-on experience in the field, and subjective recommendations were given about which tools are deemed more suitable. Given the rapid evolution of this research field, updated versions of this document might be released periodically. The authors emphasize that all software contributions to this research field are instrumental in scientific advancement and complement each other in a beneficial way.

**Table 1.** *List of Available Tools for Bayesian Networks*

| Tool | Type | ☑ Structure Learn | Param. Learn | Input Data | ☑ Infer. | Missing Data | Constraint-Based Algo. | Search/Hybrid Algo. | Scoring Fx | Param. Learn Algo. | ☑ GUI | Language | License | OS | ☑ API | Last Update | Doc. |
|---|---|---|---|---|---|---|---|---|---|---|---|---|---|---|---|---|---|
| **BayesiaLab** | Static<br>Dynamic | ☑ | ☑ | Continuous<br>Discrete<br>Mixed | ☑ | ☑ | - | MWST<br>TS<br>EQ<br>SopLEQ<br>TO | MDL | MLE | ☑ | C++ | Commercial | OSX<br>Windows<br>Linux | ☑ | 2023 | Documentation |
| **Bayes Server** | Static<br>Dynamic | ☑ | ☑ | Continuous<br>Discrete<br>Mixed | ☑ | ☑ | PC | Search & Score<br>Hierarchical<br>Chow-Liu<br>Clustering<br>TAN | LL<br>BIC | MLE | ☑ | C#<br>Java<br>Python<br>R<br>Matlab | Commercial | OSX<br>Windows<br>Linux | ☑ | 2025 | Homepage |
| **BayesFusion: GeNie (GUI) and Smile Engine** | Static<br>Dynamic | ☑ | ☑ | Continuous<br>Discrete<br>Hybrid | ☑ | ☐ | PC | ANB<br>GTT<br>Bayesian Search<br>TAN | AIC<br>B<br>Entropy<br>MDL<br>BDeu | EM<br>MLE<br>MAP | ☑ | C++ | Both | Windows | ☑ | 2025 | Documentation |
| **blip** | Static | ☑ | ☑ | Discrete | ☐ | ☑ | - | WINASOBS<br>k-MAX<br>SEM-kMAX | BIC<br>BDeu | MLE<br>BMA | ☐ | Java<br>R | Free | OSX<br>Windows<br>Linux | ☑ | 2019 | Repository |
| **BNC-Weka** | Static | ☐ | ☑ | Discrete | ☐ | ☐ | ICS | K2<br>HC<br>TAN<br>TS<br>Repeated HC<br>SA<br>Gen. Search | AIC<br>BDe<br>K2<br>MDL<br>Entropy | MLE<br>BMA | ☑ | Java | Free | OSX<br>Windows<br>Linux | ☑ | 2022 | Documentation |
| **bnlearn (Python)** | Static | ☑ | ☑ | Discrete | ☑ | ☑ | PC | TAN<br>HC<br>Chow-Liu<br>Ex. Search | BIC<br>K2<br>BDeu<br>BDs | MLE<br>BPE | ☐ | Python | Free | OSX<br>Windows<br>Linux | ☐ | 2022 | Documentation |
| **bnlearn (R)** | Static | ☑ | ☑ | Continuous<br>Discrete<br>Hybrid | ☑ | ☑ | PC-Stable<br>GS<br>IAMB<br>Fast-IAMB<br>Inter-IAMB<br>IAMB-FDR<br>Semi-Interleaved-HITON-PC<br>MPC | HC<br>ARACNE<br>TAN<br>TS<br>Chow-Liu<br>MMHC<br>HPC<br>RSMAX2 | LL<br>Predictive LL<br>K2<br>AIC<br>BIC<br>BDeu<br>BDs<br>BDJ<br>BDla<br>fNML<br>qNML<br>BGe | MLE<br>BPE<br>HEM | ☐ | R | Free | OSX<br>Windows<br>Linux | ☐ | 2025 | Homepage |
| **bnstruct** | Static | ☑ | ☑ | Continuous<br>Discrete | ☑ | ☑ | - | MMHC<br>MPC<br>HC<br>SEM<br>Silander-Myllymaki CS | AIC<br>BIC<br>BDeu | - | ☐ | R | Free | OSX<br>Windows<br>Linux | ☐ | 2024 | Homepage |
| **causal-learn** | Static | ☑ | ☐ | | ☐ | ☑ | PC<br>FCI<br>CD-NOD | GES<br>Ex. Search | BIC<br>BDeu | - | ☐ | Python | Free | OSX<br>Windows<br>Linux | ☐ | 2022 | Documentation |
| **CDT** | Static | ☑ | ☐ | Continuous<br>Discrete<br>Hybrid | ☐ | ☐ | Fast-IAMB<br>GS<br>IAMB<br>Inter-IAMB<br>MPC<br>PC | ANM<br>Bivariate Fit<br>CAM<br>CCDr<br>CDS<br>CGNN<br>GIES<br>GNN<br>GES<br>IGCI<br>Jarfo<br>LiNGAM<br>NCC<br>RCC<br>RECI<br>SAM<br>SAMv1 | BDe<br>BIC<br>GaussL0penIntScore<br>GaussL0penObsScore<br>K2<br>MDL<br>SEMGAM<br>SEMLIN<br>BDs | - | ☐ | Python | Free | OSX<br>Windows<br>Linux | ☐ | 2023 | Documentation |
| **CTBNLab** | Continuous-Time | ☑ | ☑ | | ☐ | ☐ | CT-PC<br>MB-CT-PC | HC<br>Random-restart HC<br>TS | BDe<br>CLL<br>LL | MLE<br>BPE | ☑ | Java | Free | OSX<br>Windows<br>Linux | ☐ | 2022 | Repository |

| Tool | Type | ☑ Structure Learn | Param. Learn | Input Data | ☑ Infer. | Missing Data | Constraint-Based Algo. | Search/Hybrid Algo. | Scoring Fx | Param. Learn Algo. | ☑ GUI | Language | License | OS | ☑ API | Last Update | Doc. |
|---|---|---|---|---|---|---|---|---|---|---|---|---|---|---|---|---|---|
| **DEAL** | Static | ☑ | ☑ | Continuous<br>Discrete | ☐ | ☐ | - | GES | BDe<br>BF | BPE<br>MPP | ☑ | R | Free | OSX<br>Windows<br>Linux | ☐ | 2018 | Repository |
| **gCastle** | Static | ☑ | ☐ | Continuous<br>Discrete | ☐ | ☐ | PC | ANM<br>CORL<br>DAG-GNN<br>DirectLiNGAM<br>GAE<br>GOLEM<br>GraNDAG<br>GES<br>HPCI<br>ICALiNGAM<br>MCSL<br>NOTEARS<br>NOTEARS-MLP<br>NOTEARS-SOB<br>NOTEARS-lOW-RANK<br>PNL<br>RL<br>TTPM | LL<br>AIC<br>BIC<br>LSL | - | ☐ | Python | Free | OSX<br>Windows<br>Linux | ☐ | 2022 | Repository |
| **GOBNILP** | Static | ☑ | ☐ | Discrete | ☐ | ☐ | | ILP | BDeu<br>BGe | | ☐ | C++ | Free | OSX<br>Windows | ☑ | 2018 | Homepage |
| **Hugin expert** | Static<br>Dynamic | ☑ | ☑ | Continuous<br>Discrete<br>Mixed | ☑ | ☑ | NPC<br>PC | TAN<br>Chow-Liu<br>Rebane-Pearl | AIC<br>BIC | MLE | ☑ | Java | Both | OSX<br>Windows<br>Linux | ☑ | 2019 | Homepage |
| **LiNGAM** | Static | ☑ | ☐ | Continuous | ☐ | ☐ | - | BottomUpParceLiNGAM<br>CAM-UV<br>DirectLiNGAM<br>LiM<br>LiNA<br>Longitudinal LiNGAM<br>MultiGroupDirectLiNGAM<br>MultiGroupRCD<br>RCD<br>RESIT<br>VAR-LiNGAM<br>VARMA-LiNGAM | - | - | ☐ | Python | Free | OSX<br>Windows<br>Linux | ☑ | 2023 | Documentation |
| **pcalg** | Static | ☑ | ☐ | Continuous<br>Discrete<br>Hybrid | ☐ | ☐ | FCI<br>FCI-JCI<br>Anytime FCI<br>Adaptative Anytime FCI<br>FCI+<br>PC<br>CPC<br>PC Select (PC simple)<br>RFCI | ARGES<br>GIES<br>GES<br>LiNGAM<br>Silander-Myllymaki CS | - | - | ☐ | R | Free | OSX<br>Windows<br>Linux | ☐ | 2022 | Documentation |
| **Pgmpy** | Static<br>Dynamic | ☑ | ☑ | Continuous<br>Discrete | ☑ | ☑ | PC | GES<br>HC<br>Tree Search<br>Expert in the Loop<br>MMHC<br>Ex. Search | BIC<br>K2<br>BDeu | EM<br>MLE<br>BPE<br>SEM | ☐ | Python | Free | OSX<br>Windows<br>Linux | ☐ | 2022 | Documentation |
| **Pomegranate** | Static | ☑ | ☑ | Continuous<br>Discrete<br>Mixed | ☑ | ☑ | - | HC<br>Chow-Liu<br>Exact A*<br>Exact Shortest | | EM<br>MLE | ☐ | Cython | Free | OSX<br>Windows<br>Linux | ☑ | 2025 | Documentation |
| **pyAgrum** | Static<br>Dynamic<br>Continuous-Time | ☑ | ☑ | Continuous<br>Discrete<br>Mixed | ☑ | ☑ | MIIC | GHC<br>LS-TL<br>K2<br>Chow-Liu<br>NB<br>TAN | AIC<br>BD<br>BDeu<br>BIC<br>K2<br>LL | EM | ☐ | C++<br>Python | Free | OSX<br>Windows<br>Linux | ☐ | 2023 | Homepage |

| Tool | Type | ☑ Structure Learn | Param. Learn | Input Data | ☑ Infer. | Missing Data | Constraint-Based Algo. | Search/Hybrid Algo. | Scoring Fx | Param. Learn Algo. | ☑ GUI | Language | License | OS | ☑ API | Last Update | Doc. |
|---|---|---|---|---|---|---|---|---|---|---|---|---|---|---|---|---|---|
| **Tetrad** | Static | ☑ | ☑ | Continuous<br>Discrete<br>Mixed | ☐ | ☑ | PC<br>CPC<br>PC-Max<br>FCI<br>FCI-Max<br>CFCI<br>RFCI<br>FCI-IOD<br>SvarFCI<br>SvarGFCI<br>FAS<br>FASK<br>FASK-Vote<br>MGM<br>CCD | FGES<br>IMaGES<br>IMaGES-BOSS<br>FGES-FCI<br>GRaSP-FCI<br>BOSS-FCI<br>SP-FCI<br>FGES-MB<br>BOSS<br>BPC-MIMBuild<br>FOFC-MIMBuild<br>FTFC<br>SP<br>GRaSP<br>LV-Lite<br>DAGMA<br>ICALiNGAM<br>ICA LiNG-D<br>DirectLiNGAM<br>BOSS-LiNGAM<br>CStaR<br>Orientation Algorithms<br>(R3, RSkew, Skew) | BDeu<br>CG-BIC<br>M-Separation<br>BIC<br>SEM-BIC<br>EBIC<br>GIC<br>MVP<br>PP<br>Zhang-Shen Bound | MLE | ☑ | Java<br>Python<br>R | Free | OSX<br>Windows<br>Linux | ☑ | 2025 | Documentation |
| **OpenMarkov** | Static | ☑ | ☑ | Discrete | ☑ | ☑ | PC | HC | K2 | - | ☑ | Java | Free | Windows | ☑ | 2024 | Homepage |
| **UnBBayes** | Static | ☑ | ☑ | Continuous<br>Discrete | ☑ | ☑ | - | - | B<br>CBL-A<br>CBL-B<br>K2 | - | ☑ | Java | Free | OSX<br>Windows<br>Linux | ☐ | 2020 | Homepage |

**Constraint-Based Algorithms:** Conservative Fast Causal Inference Algorithm (CFCI), Conservative Peter-Clark (CPC), Continuous-Time Peter-Clark (CT-PC), Cyclic Causal Discovery (CCD), Fast Adjacency Search (FAS), Fast Adjacency Search K (FASK), Fast Adjacency Search K Vote (FASK-Vote), Fast Causal Inference - Instrumental Observable Dependence (FCI-IOD), Fast Causal Inference - Joint Causal Inference (FCI-JCI), Fast Causal Inference (FCI), Fast Causal Inference Max (FCI-Max), Fast Incremental Association Markov Blanket (Fast-IAMB), Greedy Thick Thinning (GTT), Grow-Shrink (GS), Incremental Association Markov Blanket - False Discovery Rate (IAMB-FDR), Incremental Association Markov Blanket (IAMB), Incremental Association Markov Blanket with Peter-Clark correction (IAMBnPC), Inductive Causation (IC), Integer Linear Programming (ILP), Interleaved Incremental Association Markov Blanket (Inter-IAMB), Iterative Conditional Selection (ICS), Markov Blanket-based Continuous-Time Peter-Clark (MB-CT-PC), Max-Min Parents and Children (MPC), Mixed Graphical Model (MGM), Multivariate Information-based Inductive Causation (MIIC), Non-Parametric Conditional Independence Test (NPC), Peter-Clark Algorithm (PC), Peter-Clark Max Algorithm (PC-Max), Peter-Clark Stable Algorithm (PC-Stable), Really Fast Causal Inference Algorithm (RFCI), Semi-Interleaved HITON-Peter-Clark (Semi-Interleaved HITON-PC), Structural Vector Autoregressive Fast Causal Inference (SvarFCI), Structural Vector Autoregressive Generalized Fast Causal Inference (SvarGFCI)

**Search/Hybrid Algorithms:** Adaptive Ridge Greedy Equivalence Search (ARGES), Additive Noise Model (ANM), Algorithm for the Reconstruction of Accurate Cellular Networks (ARACNe), Augmented Naïve Bayes (ANB), Bayesian Fast Causal Inference (BFCI), Bayesian Optimization for Structure Search - Fast Causal Inference (BOSS-FCI), Bayesian Optimization for Structure Search - Linear Non-Gaussian Acyclic Model (BOSS-LiNGAM), Bayesian Optimization for Structure Search (BOSS), Bayesian Predictive Causal - Mutual Information Maximization Build (BPC-MIMBuild), Bottom-Up Partitioned Causal Learning Linear Non-Gaussian Acyclic Model (BottomUpParceLiNGAM), Causal Additive Model (CAM), Causal Additive Models with Unobserved Variables (CAM-UV), Causal Discovery in the presence of Nonstationary and/or Hetergeneous Data (CD-NOD), Causal Discovery with Soft Interventions (CDS), Causal Generative Neural Networks (CGNN), Causal Order Recovery by Learning (CORL), Causal Structure and Representation Learning (CStaR), Complete Search (CS), Cyclic Causal Discovery with Regularization (CCDr), Direct Linear Non-Gaussian Acyclic Model (DirectLiNGAM), Directed Acyclic Graph Learning via Marginal Independence (DAGMA), Directed Acyclic Graph Learning with Graph Neural Networks (DAG-GNN), Equivalence Class (EQ), Exact Search (Ex. Search), Factorial Optimization for Feature Clustering - Mutual Information Maximization Build (FOFC-MIMBuild), Fast Greedy Equivalence Search - Fast Causal Inference (FGES-FCI), Fast Greedy Equivalence Search - Markov Blanket (FGES-MB), Fast Greedy Equivalence Search (FGES), Find Two Factor Clusters (FTFC), First version of Structure Agnostic Model (SAMv1), Generalized Independence Noise condition-based Method (GIN), Gradient-based Score Propagation (GRaSP), Gradient-based Score Propagation Fast Causal Inference (GRaSP-FCI), Graph Auto-Encoder (GAE), Graph Neural Networks for Directed Acyclic Graph (GraNDAG), Graph Neural Networks for Structure Learning (GNN), Graph Optimization-based Learning of Equation Models (GOLEM), Greedy Interventional Equivalence Search (GIES), Greedy Search (GES), Hawkes Process-based Conditional Independence (HPCI) (HPCI), Hill Climbing (HC), Hybrid Parents and Children (HPC ), Independent Component Analysis for Linear Non-Gaussian Acyclic Model (ICALiNGAM), Independent Component Analysis for Linear Non-Gaussian Discovery (ICA LiNG-D), Independent Multiple-Gaussian Equivalent Search (IMaGES), Independent Multiple-Gaussian Equivalent Search with Bayesian Optimization Structure Search (IMaGES-BOSS), Inferential Methods for Graph Estimation and Search (IMGES), Information-Geometric Causal Inference (IGCI), Integer Linear Programming (ILP), Joint Approximate Regression Feature Optimization (Jarfo), k-MAX (kmax), Latent Variable Lite (LV-Lite), Linear Granger Causality (LGC), Linear Mixed (LiM), Linear Non-Gaussian Acyclic Model (LiNGAM), Linear Non-Gaussian Models for Latent Factors (LiNA), Low-Rank Approximation for Non-Equivalent Transformations for Additive Noise Models (NOTEARS-lOW-RANK), Max-Min Hill Climbing (MMHC), Maximum Weight Spanning Tree (MWST), Monte Carlo Structure Learning (MCSL), Multi-Group Direct LiNGAM (MultiGroupDirectLiNGAM), Multi-Group Repetitive Causal Discovery (MultiGroupRCD), Neural Causal Criterion (NCC), Non-Equivalent Transformations for Additive Noise Models (NOTEARS), Non-Equivalent Transformations for Additive Noise Models with Multi-Layer Perceptron (NOTEARS-MLP), Non-Equivalent Transformations for Additive Noise Models with Structural Optimization Bias (NOTEARS-SOB), Probabilistic Nonlinear Learning (PNL), Regression Error-based Causal Inference (RECI), Regression with Subsequent Independence Test (RESIT), Regression-based Causal Criterion (RCC), Reinforcement Learning for Causal Discovery (RL), Repetitive Causal Discovery (RCD), Restricted Structural Maximum Algorithm 2 (RSMAX2), Silander-Myllymaki CS, SopLEQ, Sparse Variational Graphical Fast Causal Inference (SvarGFCI), Sparsest Permutation - Fast Causal Inference (SP-FCI), Sparsest Permutation (SP), Structural Equation Modeling (SEM), Structural Expectation Maximization with k-MAX (SEM-kMAX), Structure Agnostic Model (SAM), Tabu Order (TO), Tabu Search (TS), Tree-augmented Naive Bayes (TAN), Tree-Thresholded Pairwise Mutual Information (TTPM), Vector Autoregressive Models - Linear Non-Gaussian Acyclic Model (VAR-LiNGAM), Vector Autoregressive Moving Average - Linear Non-Gaussian Acyclic Model (VARMA-LiNGAM), Weighted Incremental Association-based Structure Obtention System (WINASOBS)

**Scoring Functions:** Akaike Information Criterion (AIC), Bayes Factor (BF), Bayesian Dirichlet (BD), Bayesian Dirichlet Equivalent (BDe), Bayesian Dirichlet Equivalent Uniform (BDeu), Bayesian Dirichlet Sparse (BDs), Bayesian Gaussian Equivalent  (BGe), Bayesian Information Criterion (BIC), Bayesian Score (B), Bayesian-Dirichlet equivalent uniform with an imaginary sample size "a" (BDla), Bayesian-Dirichlet with Jeffrey's Prior (BDJ), Causal Bayesian Learning - Score A (CBL-A), Causal Bayesian Learning - Score B (CBL-B), Conditional Gaussian Bayesian Information Criterion (CG-BIC), Conditional Log-Likelihood (CLL), Directional Separation (D-Separation), Entropy, Extended Bayesian Information Criterion (EBIC), Factorized Normalized Maximum Likelihood (fNML), Gaussian L0-penalized Intervention Score (GaussL0penIntScore), Gaussian L0-penalized Observational Score (GaussL0penObsScore), Generalized Information Criterion (GIC), Generalized Score with Cross Validation (GSCV), Generalized Score with Marginal Likelihood (GSMI), K2, L0 Norm Penalty (L0NP), Least Square Loss (LSL), Log-Likelihood (LL), M-Separation (M-Separation), Minimum Description Length (MDL), Mixed Variable Polynomial (MVP), Poisson Prior (PP), Predictive Log-Likelihood (Predictive LL), quasi Normalized Maximum Likelihood (qNML), Structural Equation Modeling - Generalized Additive Model (SEMGAM), Structural Equation Modeling - Linear Model (SEMLIN), Structural Equation Modeling Bayesian Information Criterion (SEM-BIC), Zhang-Shen Bound

**Parameter Learning Methods:** Bayesian Model Averaging (BMA), Bayesian Parameter Estimation (BPE), Exhaustive Search (Ex. Search), Expectation Maximization (EM), Genetic Search (Gen. Search), Hierarchical and Expectation Maximization (HEM), Master Prior Procedure (MPP), Maximum A Posterior (MAP), Maximum Likelihood Estimation (MLE), Simulated Annealing (SA), Structural Equation Model Estimators (SEM)